\documentclass{article}
\usepackage{ijcai19}

\usepackage{times}
\usepackage{soul}
\usepackage{url}
\usepackage[hidelinks]{hyperref}
\usepackage[utf8]{inputenc}
\usepackage[small]{caption}
\usepackage{graphicx}
\usepackage{amsmath}
\usepackage{booktabs}
\usepackage{algorithm}
\usepackage{algorithmic}
\usepackage{multirow}

\title{Graph-based Neural Sentence Ordering}

\author{
Yongjing Yin$^1$\thanks{Equal contribution}\and
Linfeng Song$^{2,3*}$\and
Jinsong Su$^1$\thanks{Corresponding author}\and
Jiali Zeng$^1$\and
Chulun Zhou$^1$\And
Jiebo Luo$^2$ \\
\affiliations
$^1$Xiamen University, Xiamen, China \\
$^2$University of Rochester, Rochester, NY, U.S. \\
$^3$Tencent AI Lab, Bellevue, WA, U.S.\\
\emails
\{yinyongjing, lemon, clzhou\}@stu.xmu.edu.cn,
freesunshine0316@gmail.com,
jssu@xmu.edu.cn,
jluo@cs.rochester.edu
}

\begin{document}

\maketitle

\begin{abstract}
Sentence ordering is to restore the original paragraph from a set of sentences.
It involves capturing global dependencies among sentences regardless of their input order.
In this paper, we propose a novel and flexible graph-based neural sentence ordering model, which adopts graph recurrent network \cite{Zhang:acl18} to accurately learn semantic representations of the sentences. 
Instead of assuming connections between all pairs of input sentences,
we use entities that are shared among multiple sentences to make more expressive graph representations with less noise.
Experimental results show that our proposed model outperforms the existing state-of-the-art systems on several benchmark datasets, demonstrating the effectiveness of our model.
We also conduct a thorough analysis on how entities help the performance. 
Our code is available at https://github.com/DeepLearnXMU/NSEG.git.
\end{abstract}

\section{Introduction}

Modeling the coherence in a paragraph or a long document is an important task, which contributes to both natural language generation and natural language understanding. 
Intuitively, it involves dealing with logic consistency and topic transitions.
As a subtask, sentence ordering \cite{Barzilay_CL} aims to reconstruct a coherent paragraph from an unordered set of sentences, namely paragraph.
It has been shown to benefit several tasks, 
including retrieval-based question answering \cite{Yu_ICLR18} and extractive summarization \cite{Barzilay:jair02,Galanis:coling12},
where erroneous sentence orderings may cause performance degradation.
Therefore, 
it is of great importance to study sentence reordering.

Most conventional approaches \cite{Lapata:acl03,Barzilay_CL,Guinaudeau:acl13} are rule-based or statistical ones, relying on handcrafted and sophisticated features.
However, careful designs of these features require not only high labor costs but also rich linguistic knowledge.
Thus,
it is difficult to transfer these methods to new domains or languages.
Inspired by the recent success of deep learning, neural networks have been introduced to this task, of which representative work includes window network \cite{Li:emnlp14}, neural ranking model \cite{Chen16}, hierarchical RNN-based models \cite{Gong16,Logeswaran:aaai18}, and deep attentive sentence ordering network (ATTOrderNet) \cite{Cui:emnlp18}.
Among these models, 
ATTOrderNet achieves the state-of-the-art performance with the aid of multi-head self-attention \cite{Vaswani:nips18} to learn a relatively reliable paragraph representation for subsequent sentence ordering.

Despite the best performance ATTOrderNet having exhibited so far, it still has two drawbacks.
First, 
it is based on fully-connected graph representations.
Although such representations enable the network to capture structural relationships across sentences, 
they also introduce lots of noise caused by any two semantically incoherent sentences.
Second, 
the self-attention mechanism only exploits sentence-level information 
and applies the same set of parameters to quantify the relationship between sentences.
Obviously,
it is not flexible enough to exploit extra information, such as entities, 
which have proved crucial in modeling text coherence \cite{Barzilay_CL,Elsner:acl11}.
Thus, 
we believe that it is worthy of exploring a more suitable neural network for sentence ordering.

\begin{figure}[!t]
	\centering
	\includegraphics[width=0.9\linewidth]{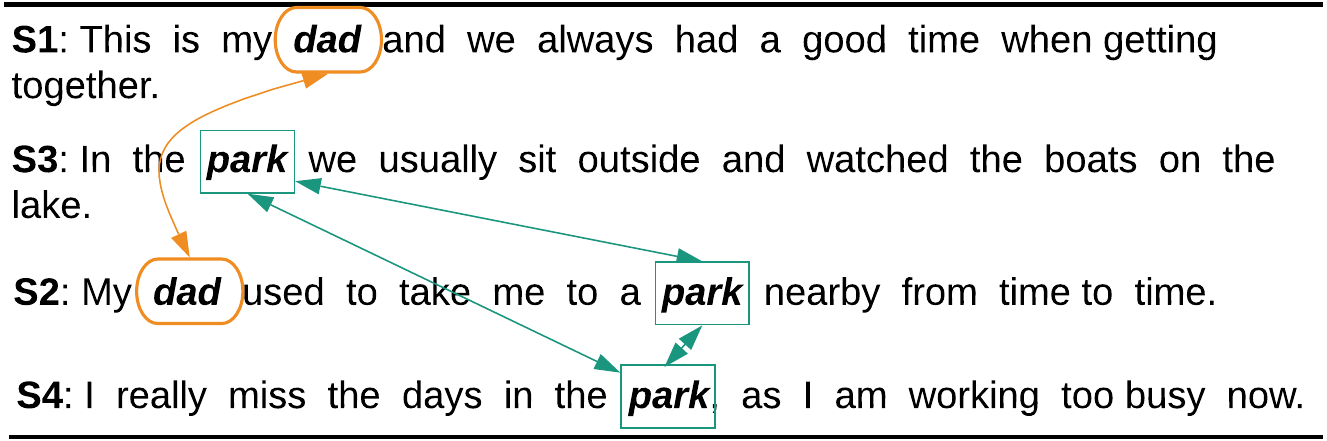}
	\caption{\label{tab:example}An example of sentence ordering, where the correct order is: \textbf{S1}, \textbf{S2}, \textbf{S3}, \textbf{S4}. \textbf{S2} is more coherent with \textbf{S1} than \textbf{S4}, as they share the same entity ``\emph{dad}''.}
	\vspace{-0.5em}
\end{figure}

In this paper, we propose a novel graph-based neural sentence ordering model
that adapts the recent graph recurrent network (GRN) \cite{Zhang:acl18}.
Inspired by Guinaudeau and Strube~\shortcite{Guinaudeau:acl13},
we first represent the input set of sentences (paragraph) as a \emph{Sentence-Entity} graph, 
where each node represents either a sentence or an entity.
Each entity node only connects to the sentence nodes that contain it, and 
two sentence nodes are linked if they contain the same entities.
By doing so,
our graph representations are able to model not only the semantic relevance between coherent sentences but also the co-occurrence between sentences and entities.
Here we take the example in Figure \ref{tab:example} to illustrate the intuition behind our representations.
We can see that the sentences sharing the same entities tend to be semantically close to each other:
both the sentences \emph{S1} and \emph{S2} contain the entity ``\emph{dad}'', and thus they are more coherent than \emph{S1} and \emph{S4}.
Compared with the fully-connected graph representations explored previously \cite{Cui:emnlp18}, 
our graph representations reduce the noise caused by the 
edges between irrelevant sentence nodes.
Another advantage is that 
the useful entity information can be fully exploited when encoding the input paragraph.
Based on sentence-entity graphs,
we then adopt GRN \cite{Zhang:acl18} to recurrently perform semantic transitions among connected nodes.
In particular,
we introduce an additional paragraph-level node to assemble semantic information of all nodes during this process,
where the resulting paragraph-level representation is beneficial to information transitions among long-distance connected nodes.
Moreover,
since sentence nodes and entity nodes play different roles, 
we employ different parameters to distinguish their impacts.
Finally, 
on the basis of the learned paragraph representation, 
a pointer network is used to produce the order of sentences.

The main contribution of our work lies in introducing GRN into sentence ordering, which can be classified into three sub aspects: 
1) We propose a GRN-based encoder for sentence ordering. Our work is the first one to explore such an encoder for this task. 
Experimental results show that our model significantly outperforms the state-of-the-arts.
2) We refine vanilla GRN by modeling sentence nodes and entity nodes with different parameters.
3) Via plenty of experiments, we verify that entities are very useful in graph representations for sentence ordering.

\section{Baseline: ATTOrderNet}
\label{sec:baseline}
\begin{figure}[!t]
	\centering
	\includegraphics[width=0.95\linewidth]{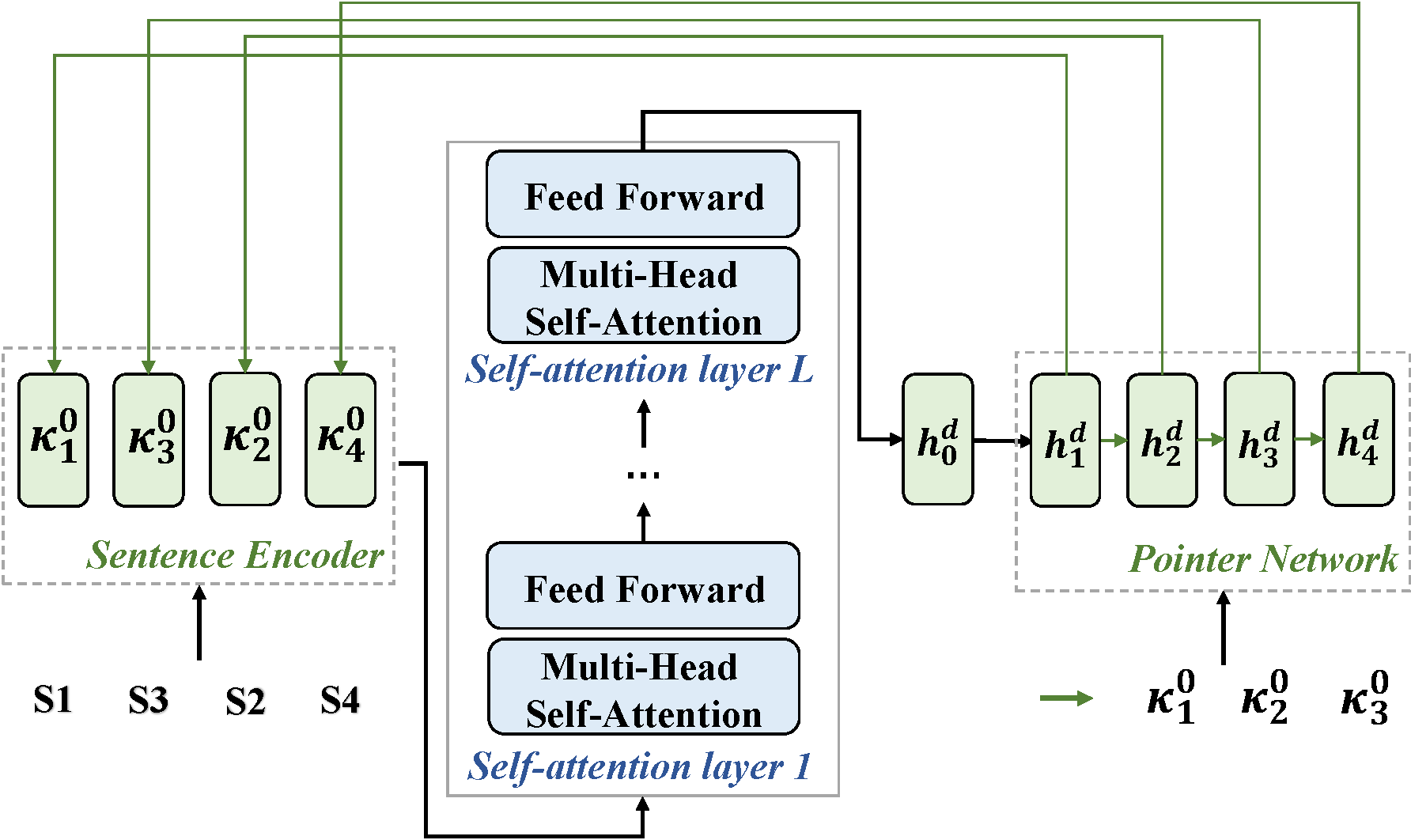}
	\caption{\label{selfatt}The architecture of ATTOrderNet.}
\end{figure}

In this section, 
we give a brief introduction to ATTOrderNet,
which achieves state-of-the-art performance and thus is chosen as the baseline of our work.
As shown in Figure \ref{selfatt}, ATTOrderNet consists of a Bi-LSTM sentence encoder, a paragraph encoder based on multi-head self-attention \cite{Vaswani:nips18}, and a pointer network based decoder \cite{Vinyals}.
It takes a set of input sentences $\boldsymbol{s}=\left[s_{o_1},\dots,s_{o_M}\right]$ with the order $\boldsymbol{o}=\left[o_1,\dots,o_M\right]$ as input and tries to recover the correct order $\boldsymbol{o}^*=\left[o_1^*,\dots,o_M^*\right]$.
Here $M$ denotes the number of the input sentences.

\subsection{Sentence Encoding with Bi-LSTM}\label{sec:base_sentencoder}
The Bi-LSTM sentence encoder takes a word embedding sequence ($\boldsymbol{x}_1,\dots,\boldsymbol{x}_{n}$) of each input sentence $s_{o_i}$ to produce its semantic representation.
At the $j$-th step, the current states ($\overrightarrow{\boldsymbol{h}}_j$ and $\overleftarrow{\boldsymbol{h}}_j$) are generated from the previous hidden states ($\overrightarrow{\boldsymbol{h}}_{j-1}$ and $\overleftarrow{\boldsymbol{h}}_{j+1}$) and the current word embedding $\boldsymbol{x}_j$ as follows:
\begin{equation}
\begin{aligned}
 \overrightarrow{\boldsymbol{h}}_j=\text{LSTM}(\overrightarrow{\boldsymbol{h}}_{j-1}, \boldsymbol{x}_j); \overleftarrow{\boldsymbol{h}}_j=\text{LSTM}(\overleftarrow{\boldsymbol{h}}_{j+1}, \boldsymbol{x}_j).
\end{aligned}
\end{equation}
Finally, the sentence representation is obtained by concatenating the last states of the Bi-LSTM in both directions $ \boldsymbol{\kappa}^0_{o_i} = [\overrightarrow{\boldsymbol{h}}_n; \overleftarrow{\boldsymbol{h}}_1]$.

\subsection{Paragraph Encoding with Multi-Head Self-Attention Network}
\label{sec:base_para_encoder}

The paragraph encoder consists of several self-attention layers followed by an average pooling layer.
Given the representations for the input sentences, the initial paragraph representation $\boldsymbol{K}^0$ is obtained by concatenating all sentence representations $\boldsymbol{K}^0=[\boldsymbol{\kappa}^0_{o_1},\dots,\boldsymbol{\kappa}^0_{o_M}]$.

Next, the initial representation is fed into $L$ self-attention layers for the update.
In particular, the update for layer $l$ is conducted by
\vspace{-0.3em}
\begin{equation} \label{eq:attn_iter}
\boldsymbol{K}^{l}=\textrm{SelfAtten}_l(\boldsymbol{K}^{l-1})\textrm{,}
\end{equation}
where $\text{SelfAtten}_l$ represents the $l$-th network layer including multi-head self-attention and feed-forward networks.
Finally, an average pooling layer is used to generate the final paragraph representation $\boldsymbol{g}$ from 
the output $K^L$ of the last self-attention layer
\label{eq:make_g}
$\boldsymbol{g} = \frac{1}{M}\sum_{i=1}^M \boldsymbol{\kappa}^L_i\textrm{,}$
where $\boldsymbol{\kappa}^L_i$ is the vector representation of $s_{o_i}$.

\subsection{Decoding with Pointer Network}
\label{sec:base_dec}
After obtaining the final paragraph representation $\boldsymbol{g}$, an LSTM-based pointer network is used to predict the correct sentence order.
Formally, the conditional probability of a predicted order $\boldsymbol{o}'$ given input paragraph $\boldsymbol{s}$ can be formalized as
\begin{equation}
\begin{aligned}
P(\boldsymbol{o}'|\boldsymbol{s}) &= \prod_{i=1}^M P(o'_i|\boldsymbol{o}'_{<i},\boldsymbol{s}) \\
P(o'_i|\boldsymbol{o}'_{<i},\boldsymbol{s})&=\textrm{softmax}(\boldsymbol{v}^T\tanh(\boldsymbol{W}\boldsymbol{h}_i^{d} + \boldsymbol{U}\boldsymbol{K}^0)),
\end{aligned}
\end{equation}
where $\boldsymbol{v}$, $\boldsymbol{W}$ and $\boldsymbol{U}$ are model parameters. 
During training, the correct sentence order $\boldsymbol{o}^*$ is known, 
so the sequence of decoder inputs 
is $\left[\boldsymbol{\kappa}^0_{o_1^*},\dots,\boldsymbol{\kappa}^0_{o_M^*}\right]$.
At test time, 
the decoder inputs correspond to the representations of sentences in the predicted order.
For each step $i$, 
the decoder state is updated recurrently by taking the representation of the previous sentence $\boldsymbol{\kappa}^0_{o'_{i-1}}$ as the input:
\begin{equation}
\boldsymbol{h}_i^{d} = \textrm{LSTM}(\boldsymbol{h}_{i-1}^{d}, \boldsymbol{\kappa}^0_{o'_{i-1}}),
\end{equation}
where $\boldsymbol{h}_i^{d}$ is the decoder state, and $\boldsymbol{h}_{0}^{d}$ 
is initialized as the final paragraph representation $\boldsymbol{g}$.
The first-step input and initial cell memory are zero vectors.

\section{Our Model}\label{sec:our_model}
In this section,
we give a detailed description to our graph-based neural sentence ordering model,
which consists of a sentence encoder, a graph neural network based paragraph encoder and a pointer network based decoder.
For fair comparison,
our sentence encoder and decoder are identical with those of ATTOrderNet.
Due to the space limitation, 
we only describe our paragraph encoder here,
which involves graph representations and graph encoding.

\begin{figure}
    \centering
    \includegraphics[width=0.6\linewidth]{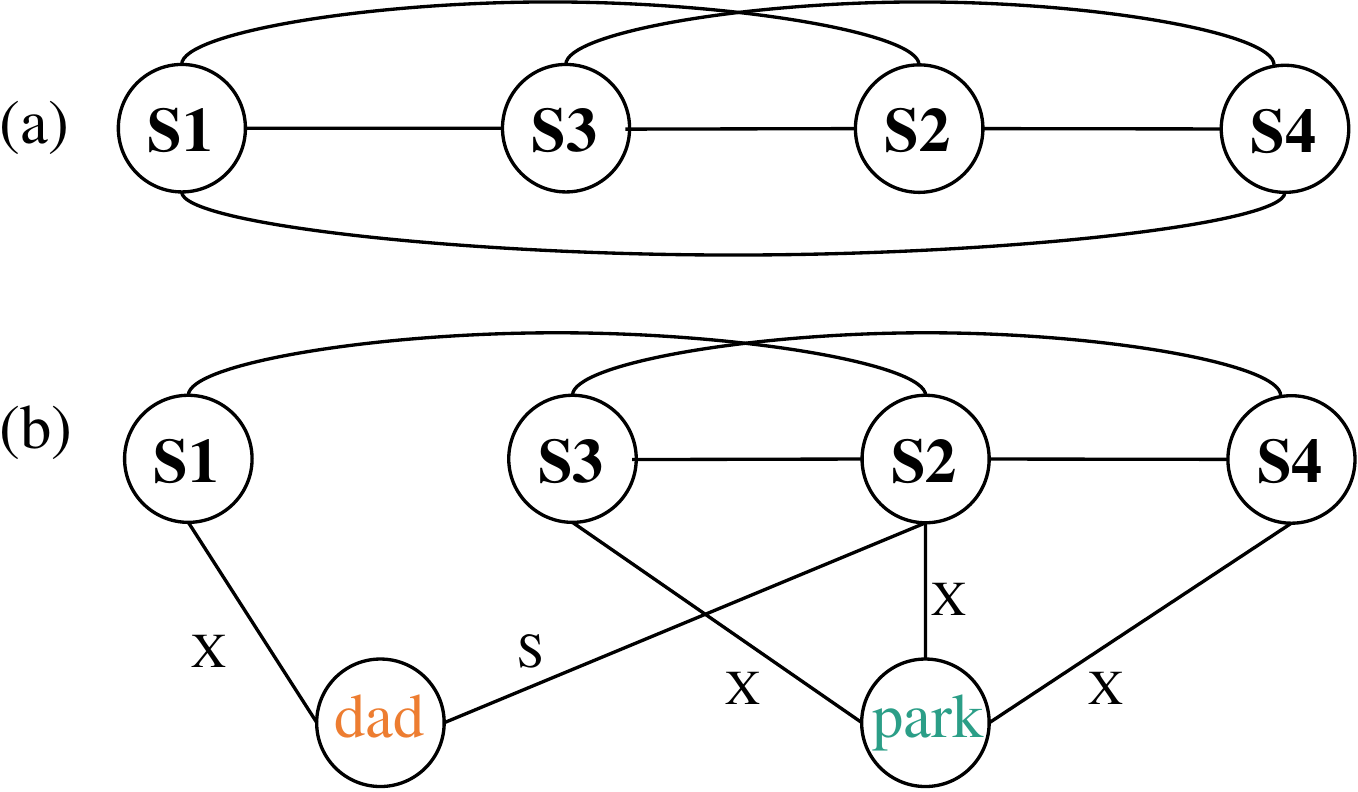}
    \caption{Comparison between (a) a fully-connected graph and (b) our sentence-entity graph for the example in Figure \ref{tab:example}. An edge label in (b) corresponds to the syntactic role of an entity in a sentence.}
    \label{fig:s_se_graphs}
\end{figure}

\subsection{Sentence-Entity Graph}
To take advantage of graph neural network for encoding paragraph,
we need to represent input paragraphs as graphs. 
Different from the fully-connected graph representations explored previously \cite{Cui:emnlp18},
we follow Guinaudeau and Strube~\shortcite{Guinaudeau:acl13} to incorporate entity information into our graphs,
where it can serve as additional knowledge and be exploited to
alleviate the noise caused by connecting incoherent sentences.
To do this, we first consider all nouns of each input paragraph as entities.
Since there can be numerous entities for very long paragraphs, 
we remove the entities that only appear once in the paragraph. 
As a result, we observe that a reasonable number of entities are generated for most paragraphs, and we will show more details in the experiments.

Then, with the identified entities,
we transform the input paragraph into a sentence-entity graph.
As shown in Figure \ref{fig:s_se_graphs} (b), our sentence-entity graphs are undirected and can be formalized as $G=(\boldsymbol{V},\boldsymbol{\hat{V}},\boldsymbol{\mathcal{E}})$, where $\boldsymbol{V}$, $\boldsymbol{\hat{V}}$ and $\boldsymbol{\mathcal{E}}$ represent the sentence-level nodes (such as $v_i$), entity-level nodes (such as $\hat{v}_j$), and edges, respectively.
Every sentence-entity graph has two types of edges, where an edge of the first type (\emph{SE}) connects a sentence and an entity within it.
Inspired by Guinaudeau and Strube~\shortcite{Guinaudeau:acl13}, we set the label for each \emph{SE}-typed edge based on the syntactic role of the entity in the sentence, which can be either a subject(\textit{S}), an object(\textit{O}) or other(\textit{X}).
If an entity appears multiple times with different roles in the same sentence, we pick the highest-rank role according to \textit{S}$\succ$\textit{O}$\succ$\textit{X}.
On the other hand, every edge of the second type (\emph{SS}) connects two sentences that have common entities, and these edges are unlabeled.
As a result, sentence nodes are connected to other sentence nodes and entity nodes, while entity nodes are only connected to sentence nodes.

Figure \ref{fig:s_se_graphs} compares a fully-connected graph with a sentence-entity graph for the example in Figure \ref{tab:example}.
Within the fully-connected graph, there are several unnecessary edges that introduce noise, such as the one connecting \emph{S1} and \emph{S4}.
Intuitively, \emph{S1} and \emph{S4} do not form a coherent context. 
It is probably because they do not have common entities, especially given the situation that they both share entities with other sentences.
In contrast, the sentence-entity graph does not take that edge, thus it does not suffer from the corresponding noise.
Another problem with the fully-connected graph is that every node is directly linked to others, 
thus no information can be obtained based on the graph structure.
Conversely, the structure of our sentence-entity graph can provide more discriminating information.



\begin{figure}
	\centering
	\includegraphics[width=1\linewidth]{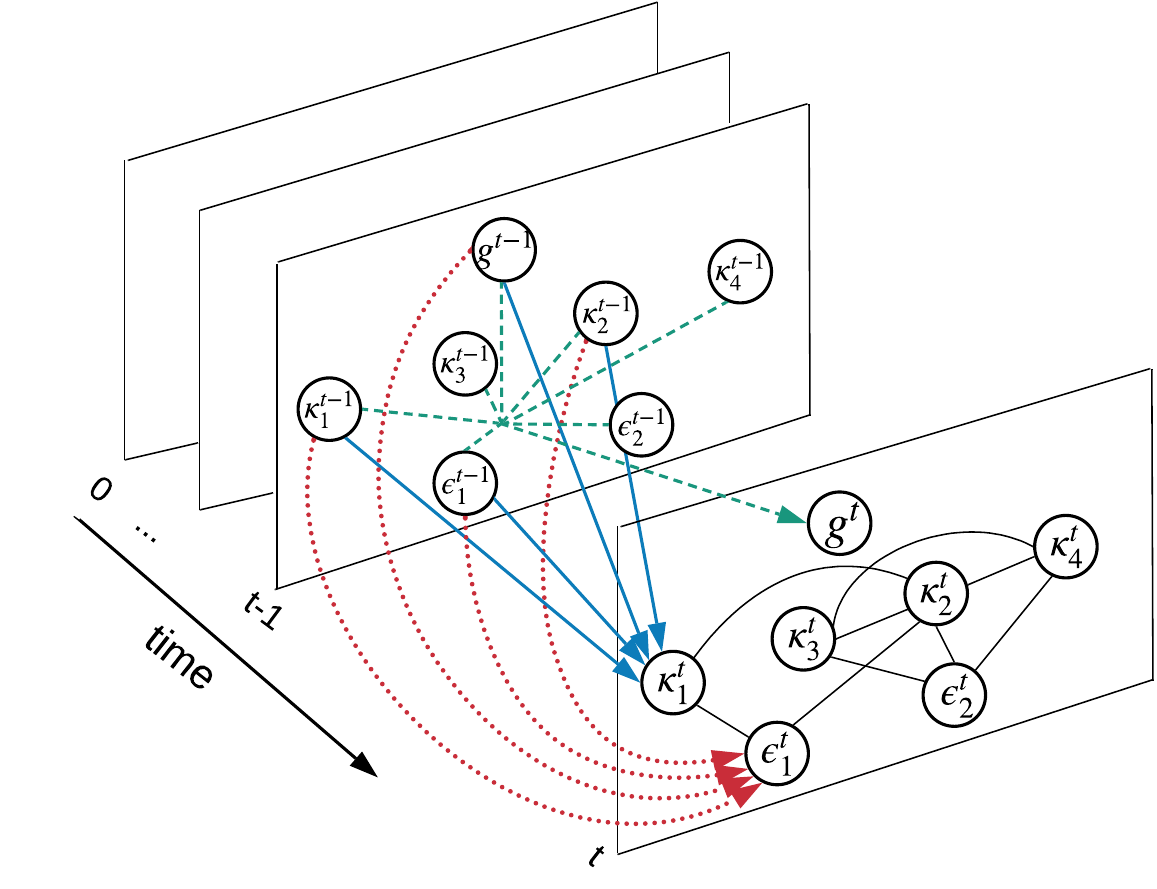}
	\vspace{-1.0em}
	\caption{\label{fig:se-graph}GRN encoding for a sentence-entity graph. The original graph structure is only drawn on step $t$ for being concise.}
\end{figure}

\subsection{Encoding with GRN}

To encode our graphs, 
we adopt GRN \cite{Zhang:acl18} that has been shown effective in various kinds of graph encoding tasks.
GRN is a kind of graph neural network \cite{scarselli2009graph} that parallelly and iteratively updates its node states with a message passing framework.
For every message passing step $t$, the state update for each node $v_i$ mainly involves two steps: a message is first calculated from its directly connected neighbors, then the node state is updated by applying the gated operations of an LSTM step with the newly calculated message.
Here we use GRU for updating node states instead of LSTM for better efficiency and fewer parameters.

Figure \ref{fig:se-graph} shows the architecture of our GRN-based paragraph encoder, which adopts a paragraph-level state $\boldsymbol{g}$ in addition to the sentence states (such as $\boldsymbol{\kappa}_i$, 
we follow Section \ref{sec:base_para_encoder} to use $\boldsymbol{\kappa}$) and entity states (such as $\boldsymbol{\epsilon}_j$).
We consider the sentence nodes and the entity nodes as different semantic units, since they contain different amount of information and have different types of neighbors.
Therefore, 
we apply separate parameters and different gated operations to model their state transition processes, both following the two-step message-passing process.
To update a sentence state $\boldsymbol{\kappa}_i^{t-1}$ at step $t$, 
messages from neighboring sentence states (such as $\boldsymbol{\kappa}_{i^\prime}^{t-1}$) and entity states (such as $\boldsymbol{\epsilon}_j^{t-1}$) are calculated via weighted sum:
\begin{equation} \label{eq:egraph_s_mess}
\begin{aligned}
    \boldsymbol{m}_i^t&=\sum_{v_{i^\prime} \in N_i} \boldsymbol{w}_{i,i^\prime}\boldsymbol{\kappa}_{i^\prime}^{t-1}\\
    \boldsymbol{\tilde{m}}_i^t&=\sum_{\hat{v}_j \in \tilde{N}_i} \boldsymbol{\tilde{w}}_{i,j,l}\boldsymbol{\epsilon}_j^{t-1}
\end{aligned}
\end{equation}
where $N_i$ and $\tilde{N}_i$ denote the sets of neighboring sentences and entities of $v_i$, respectively.
We compute gates $\boldsymbol{w}_{i,j}$ and $\boldsymbol{\tilde{w}}_{i,j,l}$ according to the edge label $l_{i,j}$ (if any) and the two associated node states using a single-layer network with a sigmoid activation. 
Then, $\boldsymbol{\kappa}_i^{t-1}$ is updated by aggregating the messages ($\boldsymbol{m}_i^t$ and$\  \boldsymbol{\tilde{m}}_i^t$) and the global state $\boldsymbol{g}^{t-1}$ via
\begin{equation} \label{eq:egraph_s_gate}
\begin{aligned}
    & \boldsymbol{\xi}_i^t=[\boldsymbol{s}_i;\boldsymbol{m}_i^t;\boldsymbol{\tilde{m}}_i^t;\boldsymbol{g}^{t-1}]\\
    & \boldsymbol{r}_i^t=\sigma\left(\boldsymbol{W}^r\boldsymbol{\xi}_i^t +\boldsymbol{U}^r\boldsymbol{\kappa}_i^{t-1}\right)\\
    & \boldsymbol{z}_i^t=\sigma\left(\boldsymbol{W}^z\boldsymbol{\xi}_i^t +\boldsymbol{U}^z\boldsymbol{\kappa}_i^{t-1}\right)\\
    & \boldsymbol{u}_i^t=tanh\left(\boldsymbol{W}^u\boldsymbol{\xi}_i^t + \boldsymbol{U}^u\left(\boldsymbol{r}_i^t \odot \boldsymbol{\kappa}_i^{t-1}\right)\right)\\
    & \boldsymbol{\kappa}_i^t=\left(\boldsymbol{1}-\boldsymbol{z}_i^t\right) \odot \boldsymbol{u}_i^t +\boldsymbol{z}_i^t \odot \boldsymbol{\kappa}_i^{t-1}
\end{aligned}
\end{equation}
Similarly, at step $t$, each entity state $\boldsymbol{\epsilon}_j^{t-1}$ is updated based on its word embedding $\boldsymbol{e}_j$, its directly connected sentence nodes (such as $\boldsymbol{\kappa}_i^{t-1}$), and the global node $\boldsymbol{g}_{t-1}$:
\begin{equation} \label{eq:egraph_e_gate}
\begin{aligned}
    \boldsymbol{\hat{m}}_j^t&=\sum_{v_i \in \hat{N}_j} \boldsymbol{\hat{w}}_{j,i,l} \boldsymbol{\kappa}_i^{t-1}\\
    \boldsymbol{\hat{\xi}}_j^t&=[\boldsymbol{e}_j;\boldsymbol{\hat{m}}_j^t;\boldsymbol{g}^{t-1}]\\
    \boldsymbol{\hat{r}}_j^t&=\sigma\left(\boldsymbol{\hat{W}}^r\boldsymbol{\hat{\xi}}_i^t +\boldsymbol{\hat{U}}^r\boldsymbol{\epsilon}_i^{t-1}\right)\\
    \boldsymbol{\hat{z}}_j^t&=\sigma\left(\boldsymbol{\hat{W}}^z\boldsymbol{\hat{\xi}}_i^t +\boldsymbol{\hat{U}}^z\boldsymbol{\epsilon}_i^{t-1}\right)\\
    \boldsymbol{\hat{u}}_j^t&=tanh\left(\boldsymbol{\hat{W}}^u\boldsymbol{\hat{\xi}}_i^t +\boldsymbol{\hat{U}}^u\left(\boldsymbol{\hat{r}}_j^t \odot \boldsymbol{\epsilon}_i^{t-1}\right)\right)\\
    \boldsymbol{\epsilon}_j^t&=\left(\boldsymbol{1}-\boldsymbol{\hat{z}}_i^t\right) \odot \boldsymbol{\hat{u}}_i^t +\boldsymbol{\hat{z}}_i^t \odot \boldsymbol{\epsilon}_i^{t-1}
\end{aligned}
\end{equation}
Finally, the global state $\boldsymbol{g}^{t-1}$ is updated with the messages from both sentence and entity states via

\begin{equation} \label{eq:egraph_g_gate}
\begin{aligned}
    \boldsymbol{\overline{\kappa}}^{t-1}&=\frac{1}{M}\sum_{m=1}^M \boldsymbol{\kappa}_m^{t-1}\\
    \boldsymbol{\overline{\epsilon}}^{t-1}&=\frac{1}{\hat{M}}\sum_{m=1}^{\hat{M}} \boldsymbol{\epsilon}_m^{t-1} \\
    \boldsymbol{r}_g^t= \sigma(\boldsymbol{W}^{sr}\boldsymbol{\overline{\kappa}}^{t-1}&+\boldsymbol{W}^{er}\boldsymbol{\overline{\epsilon}}^{t-1}+\boldsymbol{U}^{gr}\boldsymbol{g}^{t-1}) \\
    \boldsymbol{z}_g^t= \sigma(\boldsymbol{W}^{sz}\boldsymbol{\overline{\kappa}}^{t-1}&+\boldsymbol{W}^{ez}\boldsymbol{\overline{\epsilon}}^{t-1}+\boldsymbol{U}^{gz}\boldsymbol{g}^{t-1}) \\
    \boldsymbol{u}_g^t= tanh(\boldsymbol{W}^{su}\boldsymbol{\overline{\kappa}}^{t-1}&+\boldsymbol{W}^{eu}\boldsymbol{\overline{\epsilon}}^{t-1}+\boldsymbol{U}^{gu}(\boldsymbol{r}_g^t \odot \boldsymbol{g}^{t-1}))\\
    \boldsymbol{g}^t=(\boldsymbol{1}-\boldsymbol{z}_g^t) &\odot \boldsymbol{u}_g^t +\boldsymbol{z}_g^t \odot \boldsymbol{g}^{t-1},
\end{aligned}
\end{equation}
where $\boldsymbol{W}^*$ ($*$ $\in$ $\{r,z,u,sr,er,sz,ez,su,eu\}$), $\boldsymbol{U}^*$ ($*$ $\in$ $\{r,z,u,gr,gz,gu\}$), $\boldsymbol{\hat{W}}^*$ and $\boldsymbol{\hat{U}}^*$ ($* \in \{r,z,u\}$) are model parameters, and $\hat{M}$ is the number of entities.
In this way, 
each node absorbs richer contextual information through the iterative encoding process and captures the logical relationships with others. 
After recurrent state transitions of $T$ iterations, 
we obtain the final paragraph state $g^T$, which will be used to initialize the state $\boldsymbol{h}^{d}_0$ of decoder (see Section \ref{sec:base_dec}).

\begin{table*}[t] \small
	\centering
	\begin{tabular}{lcccccccccccc}  
		\toprule
		\multirow{2}{*}{Model}
		&\multicolumn{3}{c}{NIPS Abstract} &\multicolumn{3}{c}{ANN Abstract} & \multicolumn{3}{c}{arXiv Abstract} &\multicolumn{3}{c}{SIND}
		\\
		& Acc & $\tau$ & \#pm & Acc & $\tau$ & \#pm & PMR & $\tau$ & \#pm & PMR & $\tau$ & \#pm \\
		\midrule
		LSTM+PtrNet $\dagger$         & 50.87 & 0.67 & 2.1M & 58.20 & 0.69 & 3.0M & 40.44 & 0.72 & 12.7M & 12.34 & 0.48 & 3.6M \\
		V-LSTM+PtrNet $\dagger$ & 51.55 & 0.72 & 26.5M & 58.06 & 0.73 & 28.9M & -     & -    & -& -     & -    & -\\
		ATTOrderNet $\dagger$  & 56.09 & 0.72 & 8.7M & 63.24 & 0.73 & 17.9M & 42.19 & 0.73 & 23.5M & 14.01 & 0.49 & 14.4M \\
		\midrule
		F-Graph      & 56.24 & 0.72 & 4.1M & 63.45 & 0.74 & 9.9M & 42.50 & 0.74 & 19.6M & 14.48 & 0.50 & 10.6M \\
		S-Graph      & 56.67 & 0.73 & 4.1M & 64.09 & 0.76 & 9.9M & 43.37 & 0.74 & 19.6M & 15.15 & 0.50 & 10.6M \\
		SE-Graph      & \textbf{57.27}* & \textbf{0.75}* & 5.0M & \textbf{64.64}* & \textbf{0.78}* & 11.5M & \textbf{44.33}* & \textbf{0.75}* & 21.3M & \textbf{16.22}* & \textbf{0.52}* & 12.2M \\
		\bottomrule
	\end{tabular}
	\caption{Main results on the sentence ordering task, where \#pm shows the number of parameters, $\dagger$ indicates previously reported scores and * means significant at $p<0.01$ over the F-Graph on each test set. \emph{V-LSTM+PtrNet} stands for \emph{Varient-LSTM+PtrNet}. We conduct 1,000 bootstrap tests \protect\cite{koehn2004statistical} to measure the significance in metric score differences.}
	\label{tab:result}
\end{table*}

\section{Experiments}

\subsection{Setup}
\paragraph{Datasets.}
We first compare our approach with previous methods on several benchmark datasets.
\begin{itemize}
    \item \textbf{NIPS Abstract}. 
    This dataset contains roughly 3K abstracts from NIPS papers from 2005 to 2015.
    \item \textbf{ANN Abstract}. 
    It includes about 12K abstracts extracted from the papers in ACL Anthology Network (AAN) corpus \cite{Radev}.
    \item \textbf{arXiv Abstract}. 
    We further consider another source of abstracts collected from arXiv. 
    It consists of around 1.1M instances.
    \item \textbf{SIND}. It has 50K stories for the visual storytelling task\footnote{http://visionandlanguage.net/VIST/}, 
    which is in a different domain from the others. Here we use each story as a paragraph.
\end{itemize}
For data preprocessing, we first use NLTK to tokenize the sentences,
and then adopt \emph{Stanford Parser}\footnote{https://nlp.stanford.edu/software/lex-parser.shtml} to extract nouns with syntactic roles for the edge labels (\emph{S}, \emph{O} or \emph{X}).
For each paragraph, we treat all nouns appearing more than once in it as entities.
On average, each paragraph from NIPS Abstract, ANN Abstract, arXiv Abstract and SIND has 5.8, 4.5, 7.4 and 2.1 entities, respectively.

\paragraph{Settings.} Our settings follow Cui et al., \shortcite{Cui:emnlp18} for fair comparison. 
We use 100-dimension \emph{Glove} word embeddings\footnote{https://nlp.stanford.edu/projects/glove/}.
The hidden size of LSTM is 300 for NIPS Abstract and 512 for the others.
For our GRN encoders, The state sizes for sentence and entity nodes are set to 512 and 150, respectively.
The size of edge embeddings is set to 50. 
\emph{Adadelta} \cite{ADADELTA} is adopted as the optimizer
with $\epsilon$ = $10^{-6}$, $\rho$ = $0.95$ and initial learning rate 1.0.
For regularization term, 
we employ L2 weight decay with coefficient $10^{-5}$ and dropout with probability 0.5.
Batch size is 16 for training and beam search with size 64 is implemented for decoding.

\paragraph{Contrast Models.}
We compare our model (\textbf{\emph{SE-Graph}}) with the existing state of the arts, 
including 
(1) \textbf{\emph{LSTM+PtrNet}} \cite{Gong16},
(2) \textbf{\emph{Varient-LSTM+PtrNet}} \cite{Logeswaran:aaai18},
and (3) \textbf{\emph{ATTOrderNet}} \cite{Cui:emnlp18}.
Their major difference is how to encode paragraphs:
\emph{LSTM+PtrNet} uses a conventional LSTM to learn paragraph representation,
\emph{Varient-LSTM+PtrNet} is based on a set-to-sequence framework \cite{vinyals2015order},
and \emph{ATTOrderNet} adopts self-attention mechanism.
Besides, in order to better study the different effects of entities,
we also list the performances of two variants of our model:
(1) \textbf{\emph{F-Graph}}.
Similar to \emph{ATTOrderNet}, it uses a fully-connected graph to represent the input unordered paragraph, but adopts GRN rather than self-attention layers to encode the graphs.
(2) \textbf{\emph{S-Graph}}.
It is a simplified version of our model by removing all entity nodes and their related edges from the original sentence-entity graphs.
Correspondingly, all entity states ($\boldsymbol{\epsilon}$s in Equations \ref{eq:egraph_s_mess}, \ref{eq:egraph_s_gate}, \ref{eq:egraph_e_gate} and \ref{eq:egraph_g_gate}) are also removed.

\paragraph{Evaluation Metrics.}
Following previous work, 
we use the following three major metrics:
\begin{itemize}
    \item \textbf{Kendall’s tau ($\tau$}):
    It ranges from -1 (the worst) to 1 (the best).
    Specifically,
    it is calculated as 1- 2$\times$(\emph{number of inversions})/$\binom{M}{2}$,
    where $M$ denote the sequence length 
    and \emph{number of inversions} is the number of pairs in the predicted sequence with incorrect relative order.
    \item \textbf{Accuracy (Acc)}: 
    It measures the percentage of sentences whose absolute positions are correctly predicted. 
    Compared with $\tau$, 
    it penalizes results that correctly preserve most relative orders but with a slight shift.
    \item \textbf{Perfect Match Ratio (PMR)}: 
    It considers each paragraph as a single unit and calculates the ratio of exactly matching orders, 
    so no partial credit is given for any incorrect permutations.
\end{itemize}
Obviously, 
these three metrics evaluate the quality of sentence ordering from different aspects, and thus their combination can give us a comprehensive evaluation on this task.

\subsection{Effect of Recurrent Step $t$}
\begin{figure}[!t]
	\centering
	\includegraphics[width=0.8\linewidth]{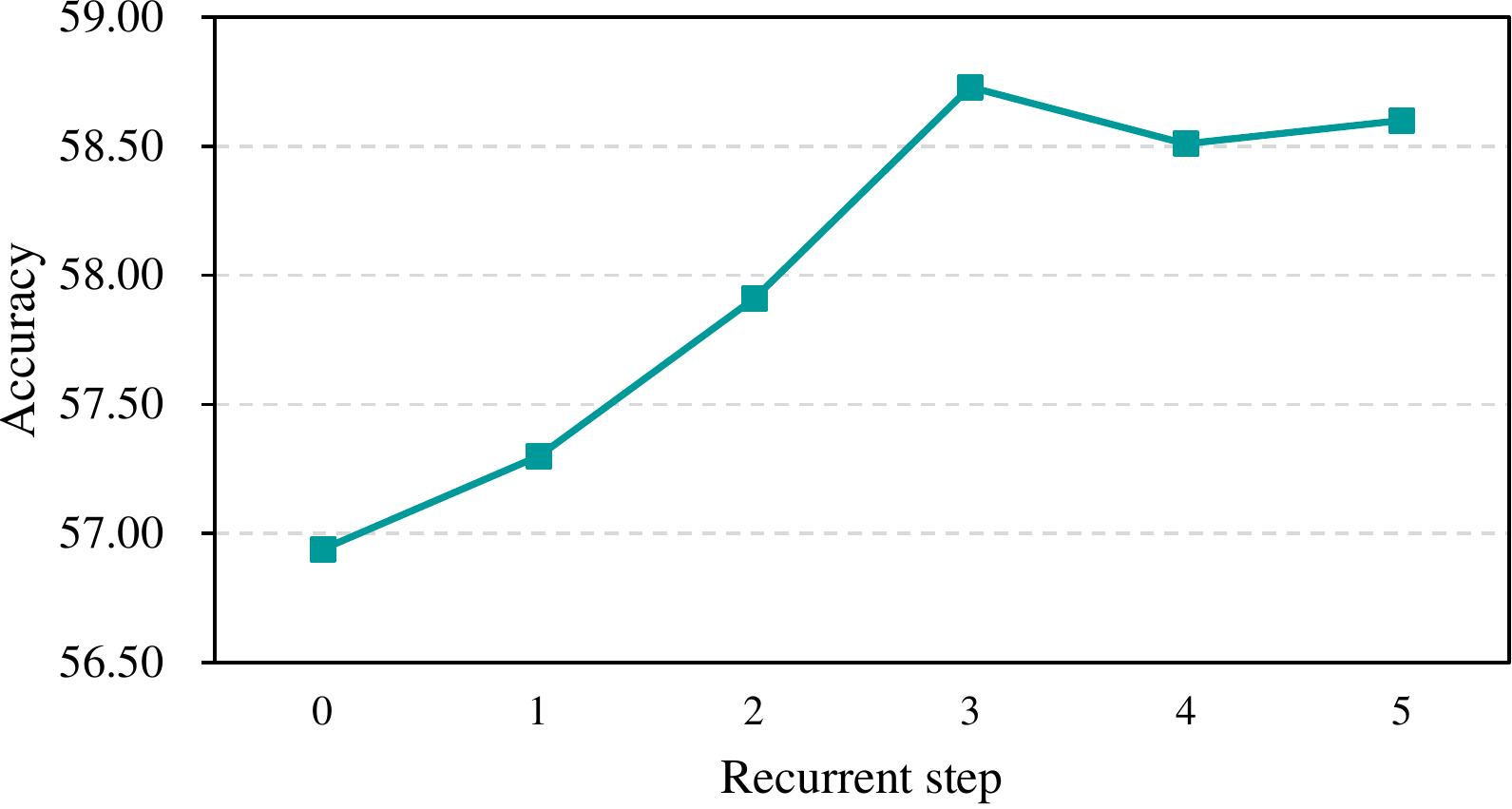}
	\caption{\label{fg:dev}Results on the arXiv Abstract validation set regarding the recurrent steps $t$.}
\end{figure}

The recurrent step $t$ is an important hyperparameter to our model, 
thus we choose the validation set of our largest dataset (arXiv Abstract) to study its effectiveness.
Figure \ref{fg:dev} shows the results.
We observe large improvements when increasing $t$ from 0 to 3, 
showing the effectiveness of our framework. 
Nevertheless, 
the increase of $t$ from 3 to 5 does not lead to further improvements while requiring more running time. 
Therefore, we set $t$=3 for all experiments thereafter.

\subsection{Main Results}
Table \ref{tab:result} reports the overall experimental results.
Our model exhibits the best performance across datasets in different domains, 
demonstrating the effectiveness and robustness of our model.
Moreover, 
we draw the following interesting conclusions. 
\textbf{First}, 
based on the same fully-connncted graph representations, \emph{F-Graph} slightly outperforms \emph{ATTOrderNet} on all datasets, 
even with fewer number of parameters and relatively fewer recurrent steps.
This result proves the validity of applying GRN to encode paragraphs. 
\textbf{Second}, \emph{S-Graph} shows better performance compared with \emph{F-Graph}.
This confirms the hypothesis that leveraging entity information can reduce the noise caused by connecting incoherent sentences.
\textbf{Third}, \emph{SE-Graph} outperforms \emph{S-Graph} on all datasets across all metrics.
It is because incorporating entities as extra information and modeling the co-occurrence between sentences and entities can further contribute to our neural graph model.
Considering that \emph{SE-Graph} has slightly more parameters than \emph{S-Graph}, we make further analysis in Section \ref{sec:analysis} to show that the improvement given by \emph{SE-Graph} is irrelevant to introducing new parameters.

Previous work has indicated that both the first and last sentences play special roles in a paragraph due to their crucial absolute positions, 
so we also report accuracies of our models on predicting them.
Table \ref{tab:headtail} summarizes the experimental results on arXiv Abstract and SIND, 
where \emph{SE-Graph} and its two variants also outperform \emph{ATTOrderNet}, and particularly, \emph{SE-Graph} reaches the best performance.
Again, both results witness the advantages of our model.

\begin{table}[!t] \small
	\centering
	\begin{tabular}{lcccccc}  
		\toprule
		\multirow{2}{*}{Model} 
		&\multicolumn{2}{c}{arXiv Abstract} &\multicolumn{2}{c}{SIND}
		\\
		& head & tail & head & tail  \\
		\midrule
		LSTM+PtrNet $\dagger$ & 90.47 & 66.49 & 74.66 & 53.30 \\
		ATTOrderNet $\dagger$ & 91.00 & 68.08 & 76.00 & 54.42 \\
		\midrule
		F-Graph & 91.43 & 68.56 & 76.53 & 56.02 \\
		S-Graph & 91.99 & 69.74 & 77.07 & 56.28 \\
		SE-Graph & \textbf{92.28} & \textbf{70.45} & \textbf{78.12} & \textbf{56.68} \\
		\bottomrule
	\end{tabular}
	\caption{The ratio of correctly predicting first and last sentences on arXiv Abstract and SIND. $\dagger$ indicates previously reported scores.}
	\label{tab:headtail}
\end{table}

\subsection{Ablation Study}
\label{sec:analysis}

To investigate the impacts of entities and edges on our model, 
we adopt \emph{SE-Graph} and \emph{S-Graph} for further ablation studies, 
because both of them exploit entity information.
Particularly, 
we continue to choose arXiv Abstract, the largest among our datasets, to conduct reliable analyses.
The results are shown in Table \ref{tab:ana}, and we have the following observations.

\textbf{First}, 
shuffling edges significantly hurts the performances of both \emph{S-Graph} and \emph{SE-Graph}.
The resulting PMR of \emph{S-Graph} (42.41) is still comparable with the PMR of \emph{F-Graph} (42.50 as shown in Table \ref{tab:result}).
Intuitively, 
shuffling edges can introduce a lot of noise.
These facts above indicate that fully-connected graphs are also very noisy, 
especially because \emph{F-Graph} takes the same number of parameters as \emph{S-Graph}.
Therefore we can confirm our previous statement again: the entities can help reduce noise.
\textbf{Second}, 
removing edge labels leads to less performance drops than removing or shuffling edges. 
It is likely because some labels can be automatically learned by our graph encoder.
Nevertheless, the labels still provide useful information.
\textbf{Third}, there are slight decreases for \emph{S-Graph} and \emph{SE-Graph}, if we only remove 10\% entities.
Removing entities is a way to simulate syntactic parsing noise, as our entities are obtained by the parsing results.
This indicates the robustness of our model against potential parsing accuracy drops on certain domains, such as medical and chemistry.
On the other hand, randomly removing 50\% entities causes significant performance drops. As the model size still remains unchanged, this shows the importance of introducing entities.
Particularly, the result of removing 50\% entities for \emph{SE-Graph} is slightly worse than original model of \emph{S-Graph}, demonstrating that \emph{SE-Graph}'s improvement over \emph{S-Graph} is not derived from simply introducing more parameters.
\textbf{Finally}, \emph{share parameters} illustrates the effect of making both GRNs (Equations \ref{eq:egraph_s_gate} and \ref{eq:egraph_e_gate}) to share parameters.
The result shows a drastic decrease on final performance, which is quite reasonable because entity nodes play fundamentally different roles from sentence nodes. 
Consequently, 
it is intuitive to model them separately.

\section{Related work}
\paragraph{Sentence Ordering.}
Previous work on sentence ordering mainly focused on the utilization of linguistic features via statistical models \cite{Lapata:acl03,Barzilay:naacl04,Barzilay_ACL05,Barzilay_CL,Elsner:acl11,Guinaudeau:acl13}.
Especially, the entity based models \cite{Barzilay_ACL05,Barzilay_CL,Guinaudeau:acl13} have shown the effectiveness of exploiting entities for this task.
Recently, the studies have evolved into neural network based models, 
such as window network \cite{Li:emnlp14}, neural ranking model \cite{Chen16}, hierarchical RNN-based models \cite{Gong16,Logeswaran:aaai18}, and ATTOrderNet \cite{Cui:emnlp18}.
Compared with these models, we combine the advantages of modeling entities and GRN, 
obtaining state-of-the-art performance.
Even without entity information, 
our model variant based on fully-connected graphs still shows better performance than the previous state-of-the-art model, 
indicating that GRN is a stronger alternative for this task.

\begin{table}[!t] \small
	\centering
	\begin{tabular}{lcccccc}  
		\toprule
		Model & \multicolumn{2}{c}{S-Graph} & \multicolumn{2}{c}{SE-Graph} \\
		      & Acc & PMR & Acc & PMR \\
		\midrule
		Original  & 58.06 & 43.37 & 58.91 & 44.33 \\
		~~~~~Shuffle edges         & 57.06 & 42.41 & 57.46 & 42.84 \\
		~~~~~Remove edge labels    & --- & --- & 58.51 & 43.96 \\
		~~~~~Remove 50\% entities  & 57.57 & 42.83 & 57.84 & 43.18 \\
		~~~~~Remove 10\% entities  & 57.79 & 43.26 & 58.67 & 44.17  \\
		~~~~~Share parameters      & --- & --- & 55.30 & 40.31 \\
		\bottomrule
	\end{tabular}
	\caption{Ablation study of our graph structure on arXiv Abstract, where \emph{Share Parameters} means employing the same parameters to update entity and sentence nodes.}
	\label{tab:ana}
\end{table}

\paragraph{Graph Neural Networks in NLP.}
Recently,
graph neural networks have been shown successful in the NLP community,
such as modeling semantic graphs \cite{Beck:acl18,Song:acl18,song2019semantic}, dependency trees \cite{marcheggiani-titov:2017:EMNLP2017,Bastings:emnlp17,Song:emnlp18}, knowledge graphs \cite{D18-1032} and even sentences \cite{Zhang:acl18,xu2018graph2seq}.
Particularly, Zhang et al.,~\shortcite{Zhang:acl18} proposed GRN to represent raw sentences by building a graph structure of neighboring words and a sentence-level node.
Their model exhibit satisfying performance on several classification and sequence labeling tasks.
Our work is in line with theirs for the exploration of adopting GRN on NLP tasks.
To our knowledge, 
our work is the first attempt to investigate GRN on solving a paragraph coherence problem.
\section{Conclusion}
We have presented a neural graph-based model for sentence ordering.
Specifically, 
we first introduce sentence-entity graphs to model both the semantic relevance between coherent sentences and the co-occurrence between sentences and entities.
Then, GRN is adopted on the built graphs to encode input sentences by performing semantic transitions among connected nodes.
Compared with the previous state-of-the-art model, 
ours is capable of reducing the noise brought by relationship modeling between incoherent sentences,
but also fully leveraging entity information for paragraph encoding.
Extensive experiments on several benchmark datasets prove the superiority of our model over the state-of-the-art and other baselines.


\section*{Acknowledgments}
The authors were supported by Beijing Advanced Innovation Center for Language Resources, National Natural Science Foundation of China (No. 61672440), the Fundamental Research Funds for the Central Universities (Grant No. ZK1024), and Scientific Research Project of National Language Committee of China (Grant No. YB135-49). 

\bibliographystyle{named}
\bibliography{ijcai19}

\end{document}